\documentclass{svproc}
\usepackage{times}
\usepackage{latexsym}
\usepackage{natbib}
\usepackage{tabularx}
\usepackage{graphicx}
\usepackage{setspace}
\usepackage{hyperref}
\usepackage{url}

\usepackage{multicol}
\usepackage[bottom]{footmisc}

\usepackage{microtype}

\usepackage{xcolor}

\begin{document}
\mainmatter
\title{Predicting Human Psychometric Properties Using Computational Language Models}
\author{Antonio Laverghetta Jr. \and Animesh Nighojkar \and Jamshidbek Mirzakhalov \and John Licato \\ 
  }
\authorrunning{Laverghetta Jr. et. al.}
\institute{Advancing Machine and Human Reasoning (AMHR) Lab \\
  Department of Computer Science and Engineering \\
  University of South Florida \\
  4202 E Fowler Ave \\
  Tampa FL 33620 \\
  USA \\
  \email{\{alaverghett,anighojkar,mirzakhalov,licato\}@usf.edu}
  }

\maketitle

\begin{abstract}
Transformer-based language models (LMs) continue to achieve state-of-the-art performance on natural language processing (NLP) benchmarks, including tasks designed to mimic human-inspired ``commonsense" competencies. To better understand the degree to which LMs can be said to have certain linguistic reasoning skills, researchers are beginning to adapt the tools and concepts from psychometrics. But to what extent can benefits flow in the other direction? In other words, can LMs be of use in predicting the psychometric properties of test items, when those items are given to human participants? If so, the benefit for psychometric practitioners is enormous, as it can reduce the need for multiple rounds of empirical testing. We gather responses from numerous human participants and LMs (transformer- and non-transformer-based) on a broad diagnostic test of linguistic competencies. We then use the human responses to calculate standard psychometric properties of the items in the diagnostic test, using the human responses and the LM responses separately. We then determine how well these two sets of predictions correlate. We find that transformer-based LMs predict the human psychometric data consistently well across most categories, suggesting that they can be used to gather human-like psychometric data without the need for extensive human trials. 
\end{abstract}

\section{Introduction}

The current generation of transformer-based language models (TLMs) \citep{vaswani2017attention} continues to surpass expectations, consistently achieving state-of-the-art results on many natural language processing (NLP) tasks. Transformers are a type of artificial neural network that connect text encoders and decoders without using recurrent links, as was the case in previous architectures such as Long Short Term Memory (LSTM) networks \citep{Hochreiter1997}. Instead, they rely on a computationally efficient self-attention mechanism \citep{vaswani2017attention}. Especially surprising is the remarkable performance of these models on benchmark tasks designed to assess ``commonsense'' reasoning (e.g., \citealp{Wang2018,Wang2019b}), possibly owing to their ability to encode and retrieve a surprising amount of structural knowledge \citep{Goldberg2019,Hu2020a,cui2020does}.

Understanding how TLMs reason is a complex task made more difficult by the fact that the sizes of contemporary TLMs are so large that they are effectively black boxes. As such, researchers are continually searching for new methods to understand the strengths and limitations of TLMs. One promising approach is to draw from the tools of psychometrics, which allows us to measure latent attributes like reasoning skills, even if the mechanisms giving rise to these attributes is not well understood. Although some have called for bridging the gap between psychometrics and artificial intelligence (AI) \citep{bringsjord2011psychometric,bringsjord2012psychometric,Hernandez2016,Wilcox2020}, the amount of work attempting to do so has been limited. While methods from psychometrics could certainly be useful as a diagnostic tool for AI practitioners, the remarkable performance of TLMs on reasoning tasks suggests that they might also be useful to psychometricians when designing evaluation scales. Most prior work has focused on the benefits psychometrics can bring to AI, however, and has not considered whether tools from AI can also benefit psychometrics, which is the focus of the present paper.

To illustrate how AI might be applied to psychometrics, assume that someone wishes to design a test to assess the degree to which a person possesses mastery of some cognitive skill $\mathcal{S}$. A good place to start is for a panel of experts to design a set of test items $\mathcal{I}$, such that they believe solving $\mathcal{I}$ requires $\mathcal{S}$, and can therefore be used to measure mastery of $\mathcal{S}$. A common task in psychometrics is to design measurement tools such as $\mathcal{I}$, and then to apply $\mathcal{I}$ to a large number of human participants. The data obtained from these trials can be used to estimate psychometric properties of the items in $\mathcal{I}$, such as their reliability, validity, and fairness. But establishing these properties can be prohibitively costly, requiring large numbers of human participants to answer the items in $\mathcal{I}$ and iteratively refine them. This drawback motivates our central research question: \textbf{Can TLMs be used to predict psychometric properties of test items?} Psychometrics would benefit greatly if so, as TLMs could be used in place of human participants, reducing the need for extensive human trials.

We present the first exploration into how well TLMs can be used to predict certain psychometric properties of linguistic test items. To do this, we identified a subset of items from the General Language Understanding Evaluation (GLUE) broad coverage diagnostic \citep{Wang2018}, a challenging benchmark of linguistic reasoning skills used to measure the progress of language modeling in the NLP community. We collected human responses on these items to assess simple psychometric properties, designing a novel user validation procedure to do so. We then assess the performance of 240 language models (LMs) on these diagnostic items. Our resulting analysis suggests TLMs show promise in modeling human psychometric properties in certain sub-categories of linguistic skills, thus providing fruitful directions for future work.

\section{Background in Natural Language Processing}
As our work draws heavily on models, datasets, and techniques from NLP, we will begin by briefly introducing some important concepts that will be used throughout this work. Note that this is not meant to be an exhaustive introduction to the field; the interested reader is encouraged to refer to the citations throughout this section for more details.

\subsection{Language Modeling}
In NLP, language models (LMs) are the primary tool used to perform tasks related to natural language understanding (e.g., sentiment analysis, machine translation, and so forth). All the models used throughout this work are examples of LMs. Given a sequence of words, the task of an LM is to predict which word is most likely to come next:

\begin{equation}
    P(w_{t}|w_{1:t-1})
\end{equation}

\noindent Where $w_t$ is the word to be predicted by the LM at timestep $t$, and $w_{1:t-1}$ is the prior $t-1$ words given to the LM to be used to make said prediction \citep{jurafsky2000speech}.

\begin{figure}
    \centering
    \includegraphics[width=0.7\linewidth]{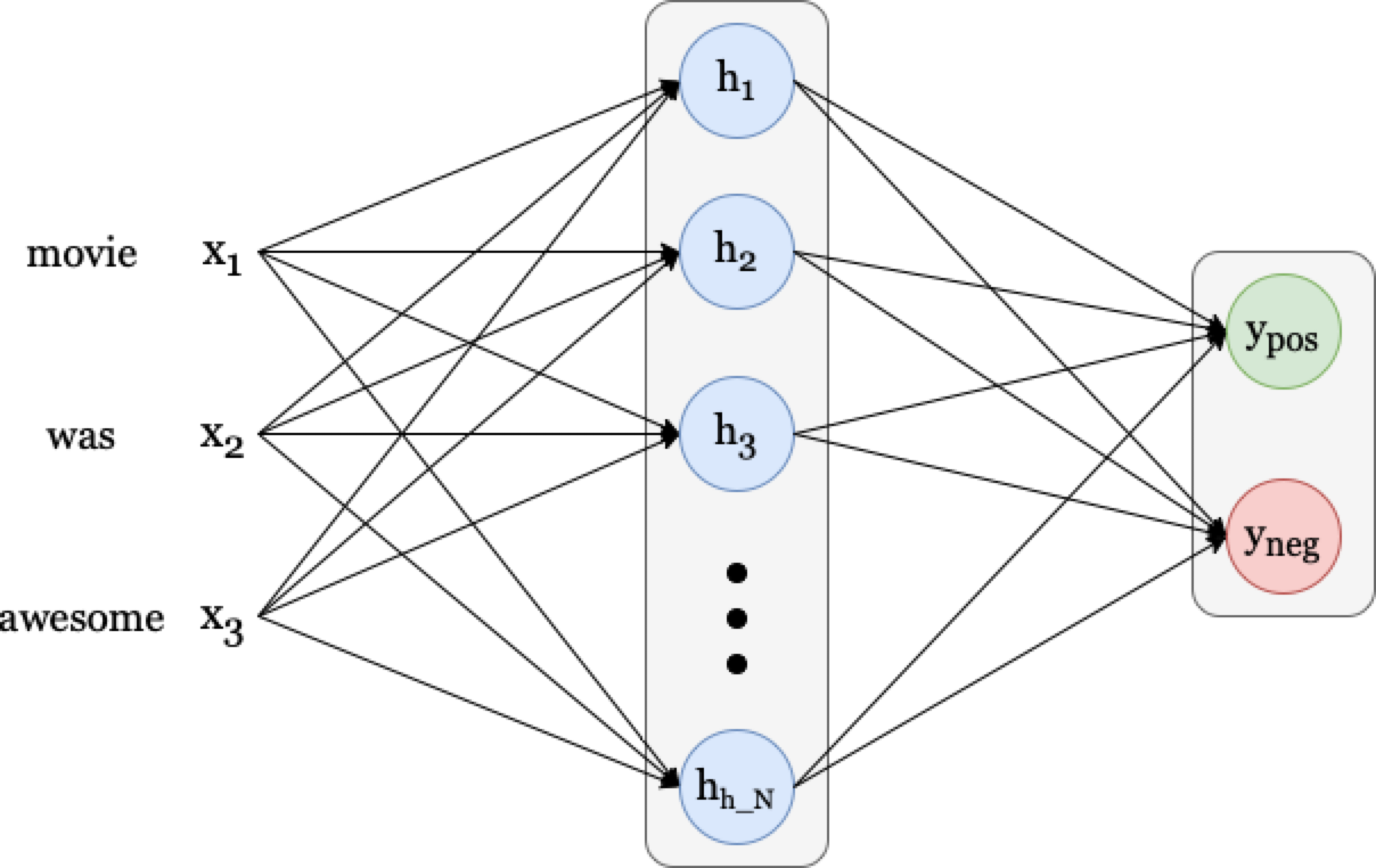}
    \caption{A simple ANN for the task of sentiment analysis. Words are input to the hidden layers, which learn to map an arbitrary sequence to a fixed output space (positive or negative sentiment). Note that the input layer is typically not counted when listing the total number of layers.}
    \label{fig:neuralLM}
\end{figure}

\noindent An LM can be constructed using a variety of probabilistic models, however, the one most relevant to this work is the artificial neural network (ANN). ANNs are models from deep learning that consist of three types of units: an input layer, one or more hidden layers, and an output layer. At a high level, they operate by taking in as input a vector representation in the input layer, performing a series of transformations on the input in each hidden layer, and finally mapping the hidden layer to fixed-length representation in the output layer. Figure \ref{fig:neuralLM} shows a schematic representation of a simple 2-layer ANN. As the hidden layers within an ANN can perform a variety of non-linear transformations to the input, ANNs are quite expressive in the kinds of representations they can learn \citep{yarotsky2021universal}, which makes them highly effective as models of language. The neural language model was first introduced in \cite{bengio2003neural}, and works by using ANNs to approximate the probability of each word, given the prior sequence of words.

\begin{figure}[t]
    \centering
    \includegraphics[width=0.7\linewidth]{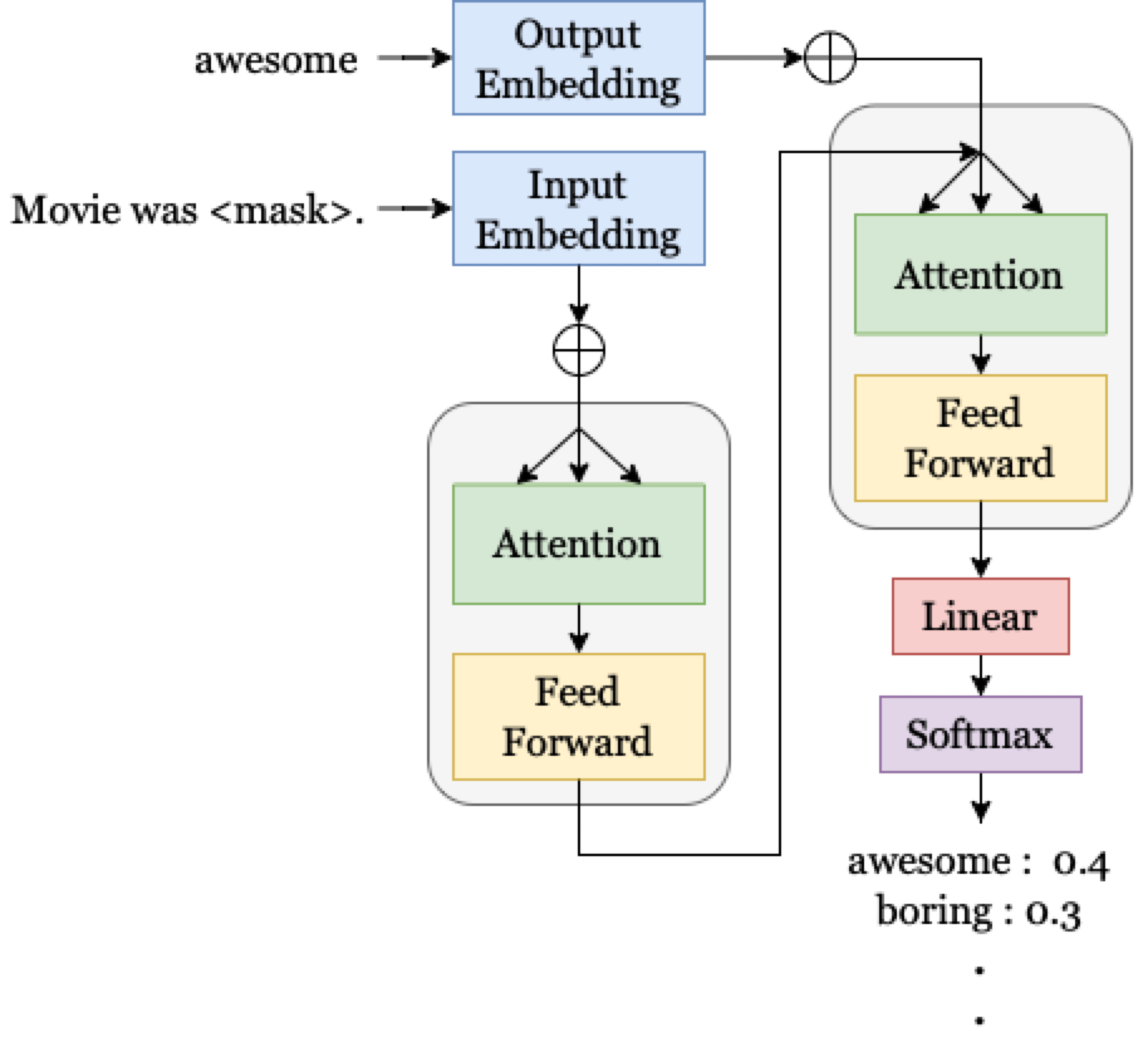}
    \caption{The architecture of the transformer. The input sequence is given with a mask token \texttt{<mask>} and the correct word is given as output. The model predicts the probability distribution for the mask token and changes its weights after comparing its predictions with the actual next word. During pre-training, this process is repeated many times over a large number of sentences and documents.}
    \label{fig:transformer}
\end{figure}

Since the advent of neural language modeling, more sophisticated neural networks have been employed in NLP, including Long Short Term Memory (LSTM) \citep{Hochreiter1997} networks, and transformers \citep{vaswani2017attention}. These types of neural networks perform the same basic set of operations as the vanilla ANN, but they differ in how their architectures are designed. LSTMs rely on using recurrent (cyclic) links between hidden layers, which allows information from previous hidden layers to affect the representations learned in later layers. Transformers rely on a technique in deep learning called attention, which is meant to mimic the attention employed in human cognitive systems. Attention works by masking out less relevant portions of the input, such that they contribute less information to later layers. For example, given the sentence ``The dog sat in the chair.", attention would learn that ``sat" and ``chair" contribute more to the meaning of ``dog" in this sentence than words like ``the." As discussed earlier, while attention has been employed in previous types of neural networks, transformers are unique in that they use only attention to learn representations of the input, throwing out recurrent layers entirely. Figure \ref{fig:transformer} shows the general structure of a transformer. A transformer block consists of only an attention operation, followed by a standard hidden layer from a typical ANN. Despite their simplicity, transformers have proven to be highly versatile models, and have surpassed the performance of previous successful architectures on virtually every NLP task.

Training any LM requires that we have access to a large corpus of text, which the LM uses to learn which words frequently occur in which contexts. While there are a variety of approaches to training an LM, by far the most successful of them was pioneered by \cite{devlin2018bert} who introduced the BERT (Bidirectional Encoder Representations from Transformers) LM. BERT is a transformer that is trained in two stages, the first being \emph{pre-training} where the model is trained using a self-supervised language modeling objective over a large corpus of text. In the second \textit{finetuning} stage the model is further trained on a labeled dataset for a particular task, thus allowing the same pre-trained model to be used for many different tasks. One can think of the pre-training stage as giving the model a large amount of domain-general knowledge, whereas the finetuning stage focuses on how to use that knowledge to solve a specific task. Most transformers introduced after BERT use this same training strategy, though the details may differ.

\subsection{Natural Language Inference}

Natural language inference (NLI) is a common task in NLP for evaluating the reasoning capabilities of LMs. NLI problems consist of two sentences: a premise ($p$) and hypothesis ($h$), and solving such a problem involves assessing whether $p$ textually entails $h$. There are typically three choices: either $p$ does textually entail $h$ (entailment), $p$ entails that $h$ is impossible (contradiction), or $h$'s truth can not be determined from $p$ alone (neutral). Whether $p$ entails or does not entail $h$ can depend on many factors, such as the syntactic relationships between the sentences, the information that the sentences convey, or some external knowledge about the world. For example, consider an NLI question with $p=$ ``My dog needs to be walked." and $h=$ ``My dogs need to be walked." We would say that $h$ contradicts $p$ because it was established in $p$ that I have only one dog. As another example, consider $p=$ ``The BART line I always take was delayed." and $h=$ ``I'm going to miss my tour of the Statue of Liberty." We might say that this is a contradiction because the BART operates in San Francisco and not New York City. However, we might also say that $p$ is neutral with respect to $h$ (perhaps I need to ride the BART to the airport, where I will then fly to New York City). Regardless, this demonstrates how the NLI task can also incorporate external information not explicitly stated in either sentence.

The NLI task was formalized in the PASCAL recognizing textual entailment tasks \citep{10.1007/11736790_9}, which were a series of workshops designed to spur the development of NLP systems for inferential reasoning. The NLI datasets developed for these tasks were quite small, having only a few thousand items in total, which made it very difficult to train deep neural networks on them. The Stanford natural language inference (SNLI) \citep{snli:emnlp2015} corpus was the first large-scale dataset of NLI questions, having around 570,000 items in total, which made it practical to train LMs for NLI. Since the release of SNLI, other large-scale NLI datasets have been curated, including MultiNLI (MNLI) \citep{N18-1101} and Adversarial NLI (ANLI) \citep{nie-etal-2020-adversarial}, each of which curates NLI questions of varying levels of difficulty and covers different domains of text (fictional stories, news, telephone conversations, etc.). This has made the NLI task quite general in the kinds of reasoning it can test for, while also being straightforward to administer to both humans and LMs, which makes the task ideal for the present study.

\subsection{Benchmarks of Commonsense Reasoning}
\label{sec:GLUE}

A common task in NLP is the development of tasks and datasets meant to assess the language understanding and reasoning capabilities of new models. Such tasks are typically narrowly scoped, focusing on how well the model performs on one specific task. More recently, there has been a trend to developing more comprehensive assessments of LM performance, meant to mimic the diverse skill sets a model would need to master when operating in the real world. The General Language Understanding Evaluation (GLUE), as well as its more recent extension SuperGLUE \citep{Wang2018,Wang2019b}, are such benchmarks and are meant to assess a broad set of linguistic reasoning competencies. GLUE was curated by combining previous datasets into a single benchmark task, covering a diverse set of underlying skills, including NLI, question answering, paraphrase detection, and others. As there has been rapid progress in NLP in recent years, the authors of GLUE found that the benchmark quickly lost the ability to discriminate between high and low-performance LMs on the tasks it covered. SuperGLUE \citep{Wang2019b} was then curated to address this, using a newer suite of more challenging tasks. 

Most relevant to this work is the GLUE task known as the broad coverage diagnostic, which is a set of items formatted as NLI problems. The diagnostic covers four main categories of linguistic competencies: \textit{lexical semantics}, \textit{predicate-argument structure}, \textit{logic}, and \textit{knowledge and common sense}. These categories are further divided into multiple sub-categories, each of which covers a specific and interesting phenomenon in language. The broad coverage diagnostic was manually curated by linguistics and NLP experts and is meant to assess broad psycholinguistic competencies of LMs across multiple categories. For instance, the \textit{propositional structure} category contains questions that exploit propositional logic operators, e.g., $p =$ ``The cat sat on the mat.'' and $h =$ ``The cat did not sit on the mat.'' The diagnostic thus aims to be a comprehensive test of linguistic reasoning skills, making it suitable for our present study. As discussed in Section \ref{sec:transformers}, we use only the following seven sub-categories from the diagnostic for our experiments:

\begin{enumerate}
    \item \textit{morphological negation}: Covers questions that require reasoning over negation in either its logical or psycholinguistic form.
    \item \textit{prepositional phrases}: Tests for the ability to handle ambiguity introduced by the insertion or removal of prepositions (e.g., $p=$ ``Cape sparrows eat seeds, along with soft plant parts and insects." and $h=$ ``Soft plant parts and insects eat seeds.").
    \item \textit{lexical entailment}: Covers hypernymy, hyponymy, and other types of monotonic relationships at the word level (e.g., a dog is an animal, but is not a cat).
    \item \textit{quantifiers}: Tests for the ability to reason over the universal and existential logical operators.
    \item \textit{propositional structure}: Tests for the ability to reason over the core suite of logical operators, including conjunction, disjunction, and conditionals.
    \item \textit{richer logical structure}: Covers higher-level forms of logic, especially those dealing with temporal or numeric reasoning.
    \item \textit{world knowledge}: Tests for knowledge of specific factual information about the world. 
\end{enumerate}

\section{Gathering Language Model Data}
\label{sec:transformers}

We begin by gathering results on the broad coverage diagnostic from a suite of LMs. We first selected a subset of the diagnostic items that were a member of only one sub-category, to better isolate factors. From this subset, we had 811 diagnostic questions encompassing 20 sub-categories. Each sub-category had at least 15 questions, and we selected the seven sub-categories enumerated in Section \ref{sec:GLUE} to use in our experiments. We selected these 7 sub-categories based on how much the average performance of the LMs improved after pre-training and finetuning. A substantial performance improvement indicated the category was solvable by the models, and would therefore provide a meaningful comparison to the human data.

We gathered responses to the diagnostic from a wide array of TLMs, including BERT \citep{devlin2018bert}, RoBERTa \citep{liu2019roberta}, T5 \citep{raffel2020exploring}, ALBERT \citep{lan2020albert}, XLNet \citep{yang2019xlnet}, ELECTRA \citep{clark2020electra}, Longformer \citep{beltagy2020longformer}, SpanBERT \citep{joshi2019spanbert}, DeBERTa \citep{he2020deberta}, and ConvBERT \citep{jiang2020convbert}. Each of these models differs from the others along one or more factors, including underlying architecture, pre-training objective and data, or the general category the model belongs to. We experimented with multiple different ``snapshots" of each TLM. We obtained these snapshots from HuggingFace \citep{wolf-etal-2020-transformers}. For each model we used a smaller version, designated with the \textit{small} or \textit{base} suffix, and a larger version, designated with the \textit{base} or \textit{large} suffix. The smaller versions of each TLM contained fewer transformer blocks, and thus fewer trainable parameters, making them less expressive models of language. We used LSTM-based LMs \citep{Hochreiter1997} as a baseline, which, unlike TLMs, primarily rely on recurrent links, as opposed to attention.

We used the SNLI \citep{snli:emnlp2015}, MNLI \citep{N18-1101}, and ANLI \citep{nie-etal-2020-adversarial} datasets to finetune our models for the NLI task. To increase the variance in our results as much as possible, we finetuned all models on various combinations of these datasets: (1) SNLI alone, (2) MNLI alone, (3) SNLI + MNLI, and (4) SNLI + MNLI + ANLI. Recall that all TLMs are trained in two stages: pre-training and then finetuning. As the performance of our models on the diagnostic will be affected by both, we systematically alter whether a model is pre-trained or finetuned to further increase variance, using the following combinations:

\begin{itemize}
    \item \textbf{Zero shot:} The model is initialized with random weights in the hidden layers and is evaluated on the diagnostic without any training. This is meant to test whether there is any property of the architecture itself which is useful for solving the diagnostic.
    \item \textbf{Pre-train, no finetune:} The model is pre-trained but not finetuned. In this case, the language model is still fully trained, but it has not been specifically optimized for NLI.
    \item \textbf{No pre-train, finetune:} The model weights are initialized randomly, but we finetune the model before evaluating it. The model is trained for NLI, but the total amount of language it has been exposed to is much smaller without pre-training.
    \item \textbf{Pre-train and finetune:} The model is fully trained before evaluation.
\end{itemize}
For BERT, we experimented with both \citet{devlin2018bert}'s pre-trained models, and a BERT model we trained from scratch. Our BERT model had an identical architecture to \textit{bert-base} and was pre-trained on Google's One Billion Words corpus \citep{chelba2014one}, which is a dataset of documents from various sources created by Google for pre-training LMs. 

In summary, this process allowed us to vary the underlying architecture, the size of each architecture, and the amount of data the model was trained on. This allowed us to treat each trained model as effectively being a different ``individual'' (and we will refer to them as such), which might have a radically different cognitive profile from its counterparts. For example, a \textit{roberta-base} model that was pre-trained and finetuned on all three NLI datasets would likely be much more proficient on our diagnostic than a \textit{roberta-large} model trained on no NLI data at all.

\section{Human Studies}
\label{sec:humans}
As our purpose in gathering this LM data was to evaluate it against human performance, we additionally ran a human study. To do this, we recruited workers on Amazon Mechanical Turk (mTurk) to complete our subset of GLUE diagnostic questions. While mTurk makes conducting large-scale human studies convenient, there are also well-documented problems with participants not completing tasks in good faith \citep{berinsky2014separating}. There are multiple techniques for filtering out bad-faith participants, such as the use of \textit{attention check} questions, sometimes called ``instructional manipulation checks'' \citep{hauser2015it}, which are designed so that a good-faith participant would be unlikely to get them incorrect. But this alone would not suffice for our purposes here, as we want a certain amount of low-scoring participants on some sub-categories, so that the population variances on sub-category items would better reflect their actual variances.

We first obtained attention checks from the ChaosNLI dataset \citep{nie2020what}, which gathered over 450,000 human annotations on questions from SNLI and MNLI. Since each question in ChaosNLI was annotated by 100 different workers, if the inter-annotator agreement for a given question is extremely high, we conclude that question is likely easy to solve for good-faith participants. We gathered 36 questions from ChaosNLI where the agreement for the correct label was at least 90\%. These were enough questions to ensure that each phase of our trials used a unique set of attention check questions. The human studies were split up into five phases, and workers who did sufficiently well in a given phase were given a qualification to continue to the next phase:

\begin{enumerate}
\item \textbf{On-boarding:} A qualifying HIT (human intelligence task) open to any worker located in the United States, who had completed at least 50 HITs with an approval rating of at least 90\%. The HIT consisted of five attention check questions, given to each worker in the same order. We gathered up to 200 responses and paid workers \$0.50.\\
\item \textbf{Phase 1:} Included questions from \textit{morphological negation}, and three attention checks. We gathered up to 45 responses and paid workers \$3.60.\\
\item \textbf{Phase 2:} Included questions from \textit{lexical entailment} and \textit{prepositional phrases}, as well as six attention checks. We gathered up to 36 responses and paid workers \$7.20.\\
\item \textbf{Phase 3:} Included questions from \textit{quantifiers} and \textit{propositional structure}, as well as six attention checks. We gathered up to 27 responses and paid workers \$7.20.\\
\item \textbf{Phase 4:} Included questions from \textit{richer logical structure} and \textit{world knowledge}, as well as six attention checks. We gathered responses from all accepted workers from Phase 3 and paid workers \$7.20.
\end{enumerate}

\noindent Our payment structure was designed to incentivize workers to put forth their best effort when completing the task. Workers were informed that successfully completing each task would award them the opportunity to earn additional payment on each subsequent phase. However, if on a given phase a worker failed our authentication protocol (described below), we rejected their work and did not pay them. Workers were informed before starting every study that we would evaluate the quality of their work, and that it might be rejected if we found evidence that they did not put forth an honest effort.

In each phase, questions were randomly ordered, except for attention checks which were spread evenly throughout the survey. We used Qualtrics\footnote{\hyperlink{https://www.qualtrics.com}{https://www.qualtrics.com}} to create the surveys for each HIT and collect the responses. Participants were first presented with instructions for the task and some examples, which were based on the instructions originally given to annotators of the MNLI dataset. The questions from each category were a randomly chosen subset of 15 questions tested on the LMs for that category, balanced for each label. For each question, workers also had to provide a short justification statement on why they believed their answer was correct, which was used to help filter out bad faith participants. To validate the responses to our surveys, we developed the following authentication procedure:

\begin{enumerate}
    \item Look for duplicate IPs or worker IDs, indicating that the worker took the HIT more than once. If there are any, keep only the first submission.\\
    \item If the worker's overall score was less than 40\%, reject the HIT. If their overall score was greater than 60\%, accept the HIT. For workers who scored between 40\% and 60\%, reject the HIT if they got less than 75\% of the attention checks correct. \\
    \item Finally, examine the justifications of all workers not previously rejected. Here we were looking for simple, but clear, reasons for why workers chose their answer. We included this step because we found in a pilot study that workers sometimes provided nonsensical justifications for their answers even when they did well on the survey, making it unclear whether they were truly paying attention. We checked that the justifications appeared relevant to the question, that they did not paste part of the question for their justification, that they did not use the same justification for every question, and that they did not use short nonsensical phrases for their justification (e.g., some simply wrote ``good'' or ``nice'' as their justification). This allowed us to keep some low-scoring participants who had put genuine effort into the task.
\end{enumerate}

\noindent Manual inspection of the resulting responses suggested that workers whose responses were accepted consistently gave higher quality responses than those who did not. These workers gave more detailed justifications that clearly articulated their thought process, often citing specific details from the question. On the other hand, workers who failed to give good justifications generally scored at or below random chance, which further indicated that they were not actually paying attention. We, therefore, believe the use of justifications helped us gather higher-quality responses. 

Using this procedure, and those described in Section \ref{sec:transformers}, we gathered results from 27 human participants and 240 neural LMs (183 transformer-based and 57 LSTM-based). In addition to the LSTMs, we also include a true random baseline which simply guesses randomly on every question. In the following experiments, we use the human performance on each category as the basis for analyzing the performance of the artificial populations, specifically using methods from classical test theory (both simple problem difficulty and inter-item correlation) and Rasch models \citep{rasch1993probabilistic} from item response theory. Our goal is to determine how well item properties measured using artificial models correlate with those measured using the humans responses, using both pearson and spearman correlation coefficients. We shall refer to the transformer population as $T$, the LSTM population as $L$, the random population as $R$, and the human population as $H$. We used the ltm R package to fit all Rasch models \citep{rizopoulos2006ltm}.

\section{Experimental Results}
\label{sec:results}

\begin{table}[htb]
\centering
\singlespacing
\caption{Given $D_H$, Spearman correlation and p-values were calculated with transformer-based ($D_T$), LSTM-based ($D_L$), and random ($D_R$) estimates of problem difficulty. Note that we have bolded cells whose correlations (absolute values) were highest, but their p-values were not always significant. Columns marked with * are significant at $p < 0.05$, ** at $p < 0.01$, and *** at $p < 0.001$.}
\begin{tabular}{r r r r}
\hline
\textbf{Category}                 & $\mathbf{D_T}$         & $\mathbf{D_L}$        & $\mathbf{D_R}$         \\
\hline
Morphological Negation   & \textbf{-0.28}     & 0.27    & -0.14              \\
Prepositional Phrases    & \textbf{***0.86} & 0.47  & 0.42  \\
Lexical Entailment       & \textbf{*0.62}  & 0.17  & -0.22  \\
Quantifiers              & \textbf{*0.57}   & -0.22 & 0.41     \\
Propositional Structure  & \textbf{***0.93} & 0.27 & 0.37  \\
Richer Logical Structure & 0.28 & -0.03  & \textbf{-0.37} \\
World Knowledge          & \textbf{***0.79} & 0.46   & -0.25\\
\hline
\end{tabular}
\label{tab:difficulties}
\end{table}

\subsection{Classical Test Theory}
We began by examining how well TLMs could predict simple problem difficulty in the human data.  For each item $i$ in a given sub-category, we calculated the percentage of human participants who got that item correct ($D^{i}_{H}$), and then the corresponding percentage for the TLMs ($D^{i}_{T}$), LSTM-based LMs ($D^{i}_{L}$), and the random baseline ($D^{i}_{R}$). We then calculated the Spearman correlation between $D^{i}_{H}$ and each of the other populations. Results are shown in Table \ref{tab:difficulties}. In almost all cases, TLMs achieve a much stronger correlation with the human data than either baseline. The main exceptions are \textit{morphological negation} and \textit{richer logical structure}, both of which fail to produce strong, statistically significant correlations. As we will see, this pattern will repeat in other measurements as well.

\paragraph{\emph{\textbf{IIC-Based Clustering}}} An important idea in psychometrics is that items that rely on the same skills should have similar chances of being answered correctly by a given participant \citep{rust2014modern}. Whether items rely on similar skills can be tested using the inter-item correlation (IIC) between two items, where high IIC suggests that the items rely on similar underlying reasoning skills. Thus, it can be assumed that if items cluster together when using IIC as a distance metric, they rely on similar underlying cognitive skills. To explore this, given a correlation measure $c$ ranging from -1 to 1, we converted it into a distance metric by taking $1-c$. We used this metric to cluster the diagnostic questions. For each sub-category, we performed clustering using human, transformer, LSTM, and random data separately ($H$, $T$, $L$, and $R$ respectively).

After clustering, for each pair of items ($i,j$) we define  $C^{D}_{i,j}$ as $1$ if $i$ and $j$ are in the same cluster as determined by dataset $D$ $\in \{H,T,L,R\}$, and $0$ otherwise. Finally, to determine how well clusters from the LM responses match the human responses, we calculated Pearson correlation between $C_H$ and each of $C_T$, $C_L$, and $C_R$. Results are shown in Table \ref{tab:clusters}. Similar to Table \ref{tab:difficulties}, we see statistically significant correlations from TLMs in every sub-category, except again for \textit{morphological negation}.

\begin{table}[htb]
\centering
\singlespacing
\caption{Pearson correlation and p-values for how well items clustered using human responses match the clusters which used transformer-based ($C_T$), LSTM-based ($C_L$), and random ($C_R$) items. Columns marked with * are significant at $p < 0.05$, ** at $p < 0.01$, and *** at $p < 0.001$.}
\begin{tabular}{r r r r}
\hline
\textbf{Category}                 & $\mathbf{C_T}$         & $\mathbf{C_L}$        & $\mathbf{C_R}$         \\
\hline
Morphological Negation   & 0.18   & \textbf{***0.40}    & -0.14    \\
Prepositional Phrases    & \textbf{**0.31}  & -0.15    & -0.01 \\
Lexical Entailment       & \textbf{**0.31} & -0.03 & -0.16    \\
Quantifiers              & \textbf{*0.24}  & -0.01 & 0.06  \\
Propositional Structure  & \textbf{***0.51} & 0.03  & 0.04  \\
Richer Logical Structure & \textbf{***0.46} & -0.07  & 0.04  \\
World Knowledge          & \textbf{**0.28} & 0.00  & -0.09 \\
\hline
\end{tabular}

\label{tab:clusters}
\end{table}

\subsection{Item Response Theory}

Since TLMs correlated well with humans using the classical techniques we tested, we wished to examine whether this would still hold using methods from item response theory (IRT). To do this, we used the diagnostic results from each population to fit Rasch models \citep{rasch1993probabilistic}. This gave us separate difficulty parameter estimates $b_i$ for each item $i$, for each population. To determine how well the difficulty parameters matched between populations, we calculated the Pearson correlation between the $b_i$ using our human response data ($H$), and the $b_i$ obtained using the other populations ($T$, $L$, $R$). Results are shown in Table \ref{tab:rasch}. As before, TLMs consistently get a stronger correlation than either baseline on most sub-categories, except for \textit{morphological negation} and \textit{richer logical structure}. Interestingly, LSTM-based LMs achieved stronger correlations than TLMs on certain sub-categories: \textit{world knowledge} and \textit{prepositional phrases}. The only other experiment where LSTM-based LMs achieved stronger correlation was reported in Table \ref{tab:clusters}, where they achieved superior correlation on \textit{morphological negation}.

\begin{table}[htb]
\centering
\singlespacing
\caption{Pearson correlation and p-values for transformer-based ($D_T$), LSTM-based ($D_L$), and random ($D_R$) estimates of problem difficulty computed using Rasch models. Columns marked with * are significant at $p < 0.05$, ** at $p < 0.01$, and *** at $p < 0.001$.}
\begin{tabular}{r r r r}
\hline
\textbf{Category}                 & $\mathbf{D_T}$         & $\mathbf{D_L}$        & $\mathbf{D_R}$         \\
\hline
Morphological Negation   & 0.08     & \textbf{0.29}    & 0.19              \\
Prepositional Phrases    & 0.48 & \textbf{**0.69}  & -0.25  \\
Lexical Entailment       & \textbf{***0.88}  & -0.06  & 0.14  \\
Quantifiers              & \textbf{*0.61}   & 0.03 & 0.12     \\
Propositional Structure  & \textbf{*0.61} & 0.05 & -0.25  \\
Richer Logical Structure & 0.16   & -0.05  & \textbf{-0.31} \\
World Knowledge          & *0.52 & \textbf{*0.59}   & -0.1  \\
\hline
\end{tabular}

\label{tab:rasch}
\end{table}

\section{Related Work}
\label{sec:related}
What reason do we have to suspect that TLMs can predict the psychometric properties of test items? Although TLMs were not primarily designed to compute in a human-like way, there are some reasons to suspect that they may have the ability to effectively model at least some aspects of human linguistic reasoning: They consistently demonstrate superior performance (at least compared to other LMs) on human-inspired linguistic benchmarks \citep{Wang2018,Wang2019b}, and they are typically pre-trained using a lengthy process designed to embed deep semantic knowledge, resulting in efficient encoding of semantic relationships \citep{zhou2020evaluating,cui2020does}. Common optimization tasks for pre-training transformers, such as the masked LM task \citep{devlin2018bert} are quite similar to the word prediction tasks that are known to predict children's performance on other linguistic skills \citep{Gambi2020}. Finally, TLMs tend to outperform other LMs in recent work modeling human reading times, eye-tracking data, and other psychological and psycholinguistic phenomena \citep{Schrimpf2020a,Schrimpf2020b,hao2020probabilistic,merkx-frank-2021-human,laverghetta-jr-etal-2021-transformer}.

Despite the potential benefits psychometrics could bring to AI, work explicitly bridging these fields has been limited. \citet{ahmad2020deep} created a deep learning architecture for extracting psychometric dimensions related to healthcare, specifically numeracy, literacy, trust, anxiety, and drug experiences. Their architecture did not use transformers and relied instead on a sophisticated combination of convolutional and recurrent layers in order to extract representations of emotions, demographics, and syntactic patterns, among others. \citet{eisape2020cloze} examined the correlation between human and LM next-word predictions and proposed a procedure for achieving more human-like cloze probabilities. In NLP, methods from IRT have been particularly popular. \citet{lalor2018understanding} used IRT models to study the impact of item difficulty on the performance of deep models on several NLP tasks. In a follow-up study, \citet{lalor2020dynamic} used IRT models to estimate the competence of LSTM \citep{Hochreiter1997} and BERT models during training. \citet{sedoc2020item} used IRT to efficiently assess chat-bots. \citet{martinez2019item} used IRT to analyze the performance of machine learning classifiers in a supervised learning task. IRT has also been used to evaluate machine translation systems \citep{otani2016irt} and speech synthesizers \citep{Oliveira2020ItemRT}. Recent work has also used IRT models to evaluate progress on benchmark NLP tasks \citep{vania-etal-2021-comparing,rodriguez-etal-2021-evaluation}. We contribute to this literature by providing what is, to our knowledge, the first comprehensive assessment of the relationships between human and LM psychometric properties on a broad test of lingusitic reasoning.
 
\section{Conclusion}
\label{sec:discuss}
Overall, we find that TLMs perform consistently better than either of our baselines in modeling human psychometric properties. However, this improvement is also not uniform across all categories. In fact, we have found some regularities in this regard. In particular, TLMs failed to achieve a strong correlation on \textit{morphological negation} in all cases. This might be explained by two facts: there is little relative variance in the human responses in this sub-category, and the average accuracy of human participants was above 90\%, as opposed to LM accuracy of 55\%.

The strong correlation TLMs consistently achieved suggests they can produce similar responses to human participants on diagnostic items. This has many implications for psychometrics, notably the possibility of using them as a sort of simulated test taker for building evaluation scales. If this were successful, it would greatly reduce the burden of multiple rounds of empirical testing.

Of course, this study also has some important limitations. The number of human participants in our study was somewhat small compared to typical psychometrics studies (which often contain hundreds or thousands of participants), making it difficult to draw stronger conclusions. As stated earlier, practical limitations on population size is a common problem in psychometrics research, one which our present work hopes to alleviate somewhat. Future work will need to repeat our experiments with much larger population sizes, and also take measures to ensure sufficient diversity in the study population (e.g., age, income, education level, English fluency, etc.). Furthermore, although we reported in detail on certain psychometrics measures where our method demonstrated promising results for TLMs, it is worth reporting that certain other measures we examined did not appear to align well. For example, item-total correlations using human data did not appear to correlate with any LM data better than with the random baseline. Likewise, our LMs failed to predict average inter-item correlations between either random subsets of items or our diagnostic sub-categories. More work is needed to better understand why.

While this study has given us some insights into which fundamental reasoning skills TLMs can model well, it does not tell us anything about the \emph{order} in which these skills are acquired, and especially whether this order is at all human-like. For example, in our experiments, we found that TLMs consistently achieved a strong correlation on items requiring mastery of logical operators and lexical entailment (e.g., $p=$ ``The dog is on the mat and the cat is in the hat" and $h=$ ``The dog is on the mat"). However, if we found that TLMs develop the ability to solve problems with conjunct-containing sentences before those with simpler sentences (e.g., $p=$ ``The dog is on the mat" and $h=$ ``The dog is not on the mat") this would clearly not reflect the order of skill acquisition we would expect to see in humans. Other methods from psychometrics, especially cognitive diagnostic models \citep{rupp2008unique} might give us a more nuanced understanding of how effective TLMs are as a model of human learning and development.

Finally, while our experiments have given us some insights into the validity and reliability of the diagnostic items, it is unclear whether our approach can allow us to measure their fairness. It is not known whether the test items we examine here are consistent across different groups of differing socio-economic statuses, and we did not control for this in our recruitment. Being able to probe this property of items would have interesting downstream applications. For instance, it might indicate whether a diagnostic gives an unfair advantage to certain types of classifiers, and thus might discriminate against certain groups.

We believe our work offers a clear path forward for bridging psychometrics and AI. The use of psychometric measures gives us a more nuanced understanding of the latent abilities of LMs than single-valued measures like accuracy or $F_1$ can provide. Furthermore, the increasingly powerful ability of TLMs to model human ``commonsense" reasoning and knowledge suggests new ways to predict psychometric properties of test items, reducing the need for costly human empirical data.

\paragraph{\emph{\textbf{Acknowledgments}}} This material is based upon work supported by the Air Force Office of Scientific Research under award numbers FA9550-17-1-0191 and FA9550-18-1-0052. Any opinions, findings, and conclusions or recommendations expressed in this material are those of the authors and do not necessarily reflect the views of the United States Air Force.

\bibliographystyle{spbasic}

\bibliography{Antonio,john,temp}

\end{document}